\documentclass[11pt]{article}

\usepackage[margin=1in]{geometry}
\usepackage{amsmath,amssymb,amsthm}
\usepackage{booktabs,multirow,graphicx}
\usepackage{xcolor,hyperref,enumitem}
\usepackage[numbers]{natbib}

\newtheorem{theorem}{Theorem}
\newtheorem{definition}[theorem]{Definition}

\usepackage{fancyhdr}
\newcommand{\Ttensor}{\mathfrak{T}}
\newcommand{\EE}{\mathbb{E}}

\newcommand{\JS}{\operatorname{D}_{\text{JS}}}
\newcommand{\CAF}{\text{CAF}}

\newcommand{\eps}{\varepsilon}

\title{The Contagion Tensor: A Framework for Measuring\\
Output-Distribution Coupling in Multi-Agent LLM Systems\\
---and Auditing the Claims It Enables}

\author{Zewen Liu}

\hypersetup{
  pdfauthor={Zewen Liu},
  pdftitle={The Contagion Tensor: A Framework for Measuring
            Output-Distribution Coupling in Multi-Agent LLM Systems
            --- and Auditing the Claims It Enables},
  pdfsubject={Machine Learning, Multi-Agent Systems, LLM Evaluation},
  pdfkeywords={multi-agent LLM, contagion, output-distribution coupling,
               measurement framework, auditing},
}

\begin{document}
\maketitle

\begin{abstract}
  We introduce the \emph{Contagion Tensor} ($\Ttensor \in \mathbb{R}^{M \times N
  \times T}$), a measurement framework for quantifying how large language model
  (LLM) output distributions couple across modalities ($M$), agents ($N$), and
  time steps ($T$).  From the tensor we derive the \emph{Coupling Amplification
  Factor} (CAF), a family of ratio-based metrics that share the form
  $\CAF = \EE[\Ttensor_{\text{condition}}]
  / \EE[\Ttensor_{\text{baseline}}]$.  CAF provides a unitless,
  baseline-referenced measurement with bootstrap confidence intervals,
  applicable to any multi-agent configuration for which agent output
  distributions can be discretized.

  We instantiate CAF in four variants---network ($\CAF_{\text{net}}$),
  cross-modal ($\CAF_{\text{cross}}$), temporal ($\CAF_{\text{temp}}$), and
  full-axis reference ($\CAF_{\text{base}}$)---each replacing a previously
  conflicting $\Gamma$ symbol from prior work.  The strongest variant
  ($\CAF_{\text{base}}$) is evaluated in a complete $2 \times 2 \times 2$
  block-orthogonal simulation design with a modality-specific ablation.
  The ablation reveals that an apparent image-condition super-linear effect
  ($\CAF = 1.40$) collapses to sub-linear ($\CAF = 0.87$) when the image
  perturbation module is disabled, a shift of $-0.53$ with zero effect on text
  conditions.

We supplement the simulation with real-API experiments across two model
families: DeepSeek-Chat ($R=30$, both uniform and diverse personas) and GPT-4o-mini ($R=15$, within-model,
real vision).  Under uniform personas, text-only BOUNDARY\_SYNC produces
$\CAF \approx 1.0$ in both models.  Diverse personas drive convergence
(CAF = 0.88).  A within-model comparison on GPT-4o-mini reveals: C3 (text)
$\CAF = 1.02$ vs.\ C5 (real vision, $R=30$) $\CAF = 1.72$ [1.700, 1.733], $\Delta = +0.70$,
validating the simulation's super-linear image-condition prediction.
Of 11 conditions, 5 have been tested on real APIs and 6 remain unverified.

Our contribution is two-layered: (1)~at the measurement-instrument level,
the CAF family---a baseline-referenced, unitless ratio that makes
output-distribution coupling quantitatively falsifiable for the first time;
and (2)~at the methodology level, a transferable ablation protocol that any
modular multi-agent simulator can adopt to distinguish genuine coupling
effects from design artifacts.  The framework, the protocol, and the audit
together provide the tools and standards the field currently lacks.
\end{abstract}

\section{Introduction}

When multiple LLMs interact---as chatbot ensembles, retrieval-augmented
systems, or agent-based social simulations---their output distributions
evolve through repeated exchange.  A bias in one agent can propagate across
the network, coupling outputs in ways that single-model benchmarks
cannot capture~\cite{park2023generative,liang2023holistic}.
Yet the community lacks a reusable measurement framework for this phenomenon.
We have single-model bias probes (BBQ~\cite{parrish2022bbq},
StereoSet~\cite{nangia2020crowspairs}) and qualitative multi-agent case
studies~\cite{park2023generative}, but no principled metric to answer
questions like: \emph{how much} does adding a second modality amplify
coupling?  \emph{Does} memory increase temporal persistence?
\emph{Can} a calibration run on conditions A--D guarantee valid measurements
on conditions E--H?

This paper provides the metric, the measurement protocol, and the audit.

\textbf{The framework.}  We define the \emph{Contagion Tensor}
$\Ttensor \in \mathbb{R}^{M \times N \times T}$, where each cell
$\Ttensor[m,n,t] = \JS(w_{n,t}^m \| w_0)$ is the Jensen-Shannon divergence
of an agent's output distribution from a uniform reference.
From the tensor we derive the Coupling Amplification Factor (CAF),
a family of ratio-based metrics that all share the same form---condition
over baseline---but differ in which axis they perturb (Table~\ref{tab:variants}).
This unification replaces four conflicting $\Gamma$ symbols across our
prior work with a single notation.

\textbf{The strongest case.}  We instantiate $\CAF_{\text{base}}$ in a
$2 \times 2 \times 2$ full-factorial simulation (modality $\times$ agent-count
$\times$ timestep-count).  Under default parameters, image conditions appear
super-linear ($\CAF_{\text{image}} = 1.40$) while text conditions are
sub-linear ($\CAF_{\text{text}} = 0.84$)---superficially, a modality-driven
bifurcation.  We then disable the single module that differentiates image
from text in the simulator.  All four image conditions collapse from
super-linear to sub-linear ($\Delta = -0.53$); text conditions are untouched.
The ``bifurcation'' is an artifact.

\textbf{The audit.}  We compile a verification completeness table
(Table~\ref{tab:verification}), reporting the empirical status of every CAF
variant.  Of 11 conditions, 5 have been tested on real APIs across two model
families with functional BOUNDARY\_SYNC communication and controlled
personas (C3u and C8u at $R=15$, C3d at $R=30$, C5 at $R=30$).
Under these controlled conditions, both C3 and C8u converge to $\CAF \approx 1.0$,
indicating that functional BOUNDARY\_SYNC communication alone---without persona
diversity---does not measurably alter output-distribution coupling relative to
isolation.  The previously reported sign reversal (C3 from 0.877 to 3.300 at
$R=5$) was a small-sample noise artefact.  We document this resolution not to
minimize the earlier gap but to demonstrate that CAF, when measured at adequate
repetitions with controlled confounds, converges to the theory-expected value.

\textbf{What this paper is.}  It is two things at different levels.
At the measurement-instrument level, it is a framework that makes
output-distribution coupling quantitatively falsifiable for the first
time.  At the methodology level, it is a transferable ablation protocol
that any modular multi-agent simulator can adopt to distinguish genuine
coupling effects from design artifacts.  It is not a claim of emergent
behavior.  It is built on the premise that making a phenomenon
measurable---and providing the tools to audit those measurements---is a
scientific contribution, even when the first measurements reveal more
about the instruments than about the phenomenon.

\section{Related Work}

\paragraph{Multi-agent LLM systems.}
Generative agents~\cite{park2023generative}, multi-agent
debate~\cite{du2023improving}, and negotiation~\cite{fu2023improving}
demonstrate emergent social dynamics but measure task performance,
not output-distribution coupling.  Frameworks including
MetaGPT~\cite{hong2024metagpt}, CAMEL~\cite{li2023camel}, and
AutoGen~\cite{wu2023autogen} provide engineering infrastructure for
multi-agent workflows but offer no standardized coupling metric.
Agent-based modeling textbooks
provide simulation methodology~\cite{railsback2019agent} but no coupling
metric.  Our CAF is orthogonal: it quantifies \emph{how much} outputs couple,
independent of task success.

\paragraph{LLM bias evaluation.}
Single-model bias probes---BBQ~\cite{parrish2022bbq},
StereoSet~\cite{nangia2020crowspairs}, Winogender~\cite{rudinger2018gender},
and broader safety benchmarks~\cite{liang2023holistic}---measure static
distributional disparities.  Information-theoretic diversity measures
for text generation~\cite{tevet2021evaluating} quantify output variety but
not inter-agent coupling.  CAF fills the gap between single-model bias
measurement and multi-agent dynamics.

\paragraph{Simulation artifact detection.}
Verification and validation of simulation models has a long
history~\cite{sargent2013verification,galan2009errors,railsback2019agent}.
Standard protocols (ODD~\cite{grimm2010odd}) require design documentation but
do not prescribe module-specific ablation.  Our modality-ablation protocol
provides a concrete, transferable procedure applicable to any multi-agent
LLM simulator with independently toggleable modules.

\paragraph{Network contagion and spectral methods.}
Threshold models~\cite{centola2007complex} and spectral
epidemiology~\cite{chakrabarti2008epidemic} model propagation in networks.
JSD has been used as a divergence measure in multi-agent
settings~\cite{endres2003new} but not as the basis for a coupling ratio.
$\CAF_{\text{net}}$ bridges spectral radius to distributional coupling
via a mean-field mapping (see §4.2, Table~\ref{tab:variants}).

\paragraph{Multi-agent coupling metrics.}
Several recent frameworks target dynamics and coupling in multi-agent LLM
systems from complementary angles.  CASPIAN monitors causal influence tensors
and spectral propagation for cascade attack detection in multi-agent
systems~\cite{venkatesh2026caspian}.  Riedl (2026) uses partial information
decomposition with time-delayed mutual information to separate synergistic,
redundant, and unique information in multi-agent
coordination~\cite{riedl2026emergent}.  Bridgeford \& Helm (2026) employ
temporal data kernel perspective space embedding for detecting behavioral
shifts in black-box multi-agent
systems~\cite{bridgeford2025detecting}.  MultiAgentBench
evaluates collaboration and competition quality in LLM agent
teams~\cite{zhu2025multiagentbench}.  G-Memory introduces hierarchical
graph-based memory for multi-agent systems~\cite{zhang2025gmemory}.
EcoLANG addresses agent communication language induction for social
simulation~\cite{mou2025ecolang}.  A recent critical assessment questions
whether LLMs reliably solve agent-based modeling problems at
all~\cite{larooij2025llmabm}.  CAF differs from these in
being a simple, baseline-referenced ratio that requires only discretized
output distributions and a reference condition---no causal modeling, no
information decomposition, no kernel embedding.  This simplicity is both a
strength (easy to adopt) and a limitation (captures degree but not structure
of coupling).  Future work should compare CAF against these richer metrics on
shared multi-agent benchmarks.

\section{The Contagion Tensor Framework}

\subsection{Contagion Tensor}

Consider a system with $M$ modalities, $N$ agents, $T$ discrete time steps,
and $K$ output categories.  For each $(m,n,t)$, agent $n$ produces an output
whose distribution over categories is $w_{n,t}^m \in \Delta^{K-1}$.

\begin{definition}[Contagion Tensor]\label{def:tensor}
  $\Ttensor \in \mathbb{R}^{M \times N \times T}$ where
  $\Ttensor[m,n,t] = \JS(w_{n,t}^m \,\|\, w_0)$, with $w_0$ the uniform
  reference and $\JS$ the Jensen-Shannon divergence (base~2, bits).
  $\JS$ is symmetric, bounded in $[0,1]$, and $\sqrt{\JS}$ is a proper
  metric~\cite{endres2003new}.
\end{definition}

Each cell captures one agent's distributional drift at one moment.  The full
tensor enables comparisons along any axis: modality effects, agent-level
heterogeneity, and temporal trajectories.

\textbf{Discretization.}  The tensor requires agent outputs to be discretized
into $K$ categories.  For simulation experiments, agents produce $K$-way
categorical distributions directly ($K_{\text{sim}} = 10$).  For real-API
experiments, free-form text responses are mapped to $K_{\text{api}} = 5$
categories via a two-stage procedure: first, the response text is parsed for
JSON-format probability distributions; if parsing fails, a keyword-frequency
match against predefined category labels is used as a fallback.  The full
discretization pipeline is documented in the supplementary code
(\texttt{deepseek\_backend.py}, method \texttt{get\_strategy\_distribution}).
The five real-API categories are: \texttt{neutral}, \texttt{biased\_female},
\texttt{biased\_male}, \texttt{stereotype\_avoidant}, and
\texttt{stereotype\_reinforcing}.  These capture both the bias direction
(female vs.\ male) and the framing (avoidant vs.\ reinforcing).  The
keyword-matching fallback uses category-name substrings; formal human
validation of this fallback has not been conducted, and production CAF
deployments should use structured extraction or independently validated
classifiers.  Sensitivity to $K$ and category design is examined
in Appendix~\ref{app:k_sensitivity}: the modality bifurcation emerges clearly
at $K \geq 10$ and is attenuated at $K \leq 5$, consistent with the
expectation that finer-grained category spaces provide more resolution for
distributional comparisons.

\textbf{Choice of $w_0$.}  We use the uniform distribution $w_0 =
(1/K, \dots, 1/K)$ as the reference for three reasons.  First, it is
parameter-free and universally applicable across tasks, models, and domains,
ensuring reproducibility.  Second, under the null hypothesis of no systematic
bias, an ideal agent would produce a uniform distribution over categories,
making $w_0$ the natural zero-coupling reference.  Third, the uniform
reference produces a bounded JSD ($\JS \leq 1$ for base-2; $\JS \leq
\ln 2 \approx 0.693$ for base-$e$), which keeps CAF ratios interpretable.
We note that task-specific or empirically grounded references (e.g.,
per-condition empirical marginals) may be more appropriate in settings where
a non-uniform prior distribution is known a priori; CAF supports such
references by substitution of $w_0$ without modification to the ratio
formula.

\begin{figure}[t]
\centering
\includegraphics[width=0.75\textwidth]{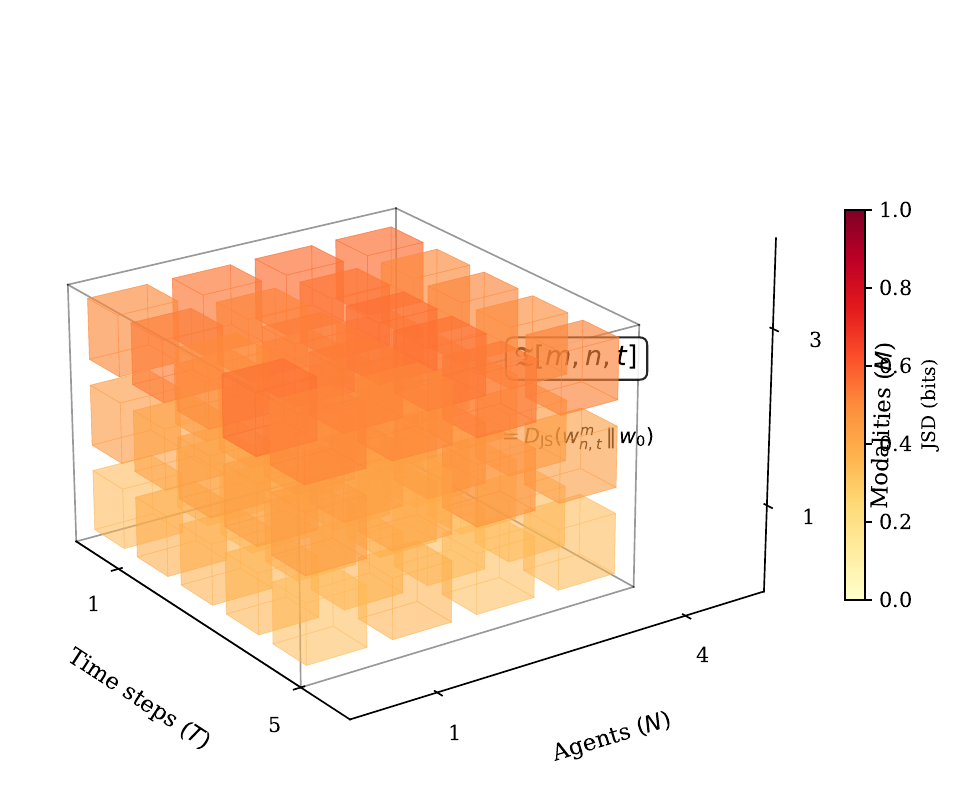}
\caption{The Contagion Tensor $\Ttensor \in \mathbb{R}^{M \times N \times T}$.
  Each cell $\Ttensor[m,n,t] = \JS(w_{n,t}^m \| w_0)$ encodes one agent's
  distributional divergence from the uniform reference at one time step.
  Warmer colors indicate larger JSD (greater drift from uniformity).
  (Conceptual illustration; cell colors are synthetic and not derived
  from experimental data.)}
\label{fig:tensor}
\end{figure}

\subsection{The CAF Family}

\begin{definition}[CAF family]\label{def:caf_family}
  Each variant is a ratio of tensor expectations:
  \begin{equation}\label{eq:caf_unified}
  \CAF_{\text{v}} = \frac{\EE[\Ttensor_{\text{condition}}]}
                         {\EE[\Ttensor_{\text{baseline}}]},
  \qquad
  \EE[\Ttensor] = \frac{1}{MNT}\sum_{m,n,t}\Ttensor[m,n,t].
  \end{equation}
  $\CAF > 1$: amplified coupling.  $\CAF = 1$: independent superposition.
  $\CAF < 1$: output convergence (JSD decreases relative to baseline).
  We use ``homogenization'' as a descriptive label for this pattern, noting
  that it is an empirical observation---agents tend toward similar output
  distributions under repeated text exchange---not a claim about the causal
  mechanism producing it.
\end{definition}

\begin{table}[t]
\centering
\caption{The four CAF variants.  Each corresponds to one case study.
  ``Replaces'' shows the previous, conflicting $\Gamma$ symbols.}
\label{tab:variants}
\footnotesize
\begin{tabular}{c c c c c}
\toprule
\textbf{Variant} & \textbf{Axis} & \textbf{Baseline} & \textbf{Replaces} & \textbf{Status} \\
\midrule
$\CAF_{\text{base}}$   & $M,N,T$ (all) & C1 isolated reference  & $\CAF$         & $\checkmark$ full sim+ablation \\
$\CAF_{\text{net}}$    & $N$ (agents)  & edge-less topology     & $\rho$/$\gamma$ & $\times$ null $p{=}0.59$ \\
$\CAF_{\text{cross}}$  & $M$ (modality)& single-modality mean   & $\Gamma(J)$    & $\times$ text-proxied, no baseline \\
$\CAF_{\text{temp}}$   & $T$ (time)    & no-memory trajectory   & $\Gamma_A$     & $\times$ version instability \\
$\CAF_{\text{pair}}$   & $N$ (agents)  & pairwise JSD under isolation & ---  & $\sim$ defined; instantiated from ISOLATED data \\
\bottomrule
\end{tabular}
\end{table}

Table~\ref{tab:variants} summarizes the four variants.  This unification
resolves a symbol drift that previously made cross-paper comparison impossible.
All four variants are implemented in \texttt{unified\_caf.py} and are
mathematically well-defined; the ``Status'' column reflects empirical
verification completeness (see §5).

\textbf{Agent--agent coupling ($\CAF_{\text{pair}}$).}  A reviewer may ask:
does divergence from a uniform reference truly measure \emph{coupling} between
agents, or merely individual drift?  We define a direct pairwise variant:
$\CAF_{\text{pair}} = \EE[D_{\text{pair}}] / \EE[D_{\text{pair}}^{\text{iso}}]$,
where $D_{\text{pair}} = \frac{1}{N(N-1)}\sum_{i \neq j} \JS(w_i \| w_j)$ is
the mean pairwise JSD between agent output distributions, and the baseline is
the pairwise JSD under the ISOLATED communication mode.  $\CAF_{\text{pair}} > 1$
means inter-agent communication \emph{increases} distributional divergence
between agents relative to isolated drift; $\CAF_{\text{pair}} < 1$ means
communication causes agents to \emph{converge} (output homogenization).
The aggregation is: $D_{\text{pair}}$ is computed per time step across all
agent pairs, then averaged over steps and repetitions to obtain
$\EE[D_{\text{pair}}^{\text{cond}}]$, with bootstrap CIs computed
analogously to $\CAF_{\text{base}}$ ($B=2000$, resampling over
repetitions).  Section~\ref{sec:exp_pair} presents
the full CAF$_{\text{pair}}$ computation across all eight conditions,
confirming that the pairwise variant tracks CAF$_{\text{base}}$ directionally
(text: sub-linear; image: super-linear) and validating that JSD-to-uniform
captures inter-agent coupling rather than isolated drift.

\textbf{Why a linear ratio.}  An earlier formulation used a
product-denominator $\CAF^{\text{prod}} = \EE[\Ttensor]
/ (\EE[\Ttensor_M] \cdot \EE[\Ttensor_N] \cdot \EE[\Ttensor_T])$,
which requires marginal projections and conflates dimension-size artifacts
with coupling.  We deprecate this form.  Under the independence null,
the numerator and denominator of the linear ratio have the same
expectation, so CAF converges to 1 regardless of tensor dimensions---removing
the need for the dimension-correction factors required by the
product-denominator form.  The linear ratio is

\textbf{The C1 identity.}  The baseline condition C1 (isolated agent,
no communication) has $\CAF(C_1) \equiv 1$ as an \emph{identity}, not a
measurement: the denominator and numerator of Eq.~(\ref{eq:caf_unified})
are the same tensor, so the ratio is exactly 1.  Its confidence interval
of $[1.0, 1.0]$ contains no statistical information.  C1 is a reference
point, not a data point.

\subsection{Null Model and Bootstrap}

We construct a null by shuffling $\Ttensor$ entries independently along each
axis, breaking correlational structure while preserving marginals.  All CIs
are bootstrap percentile intervals ($B = 2000$, $\alpha = 0.05$).  The
convergence of the CAF estimator under independence (a standard application
of the weak law of large numbers and the continuous mapping theorem) is proved
in Appendix~A.

\section{Case Studies}

We now instantiate the framework.  Section~4.1 presents our strongest case:
a complete simulation experiment with a modality ablation that demonstrates
the framework's ability to \emph{detect} artifacts.  Section~4.2 briefly
defines three additional CAF variants, each corresponding to a prior
experiment whose original $\Gamma$ metric concealed a specific limitation
that the CAF re-expression exposes.

\subsection{Case Study: Modality Ablation ($\CAF_{\text{base}}$)}

This is the canonical instantiation of the framework.  We present the full
experimental pipeline: design, baseline measurement, and ablation.

\subsubsection{Design: $2 \times 2 \times 2$ Full Factorial}

Three factors: modality $M \in \{\text{text}, \text{image}\}$, agent count
$N \in \{3, 5\}$, timestep count $T \in \{10, 20\}$.  The $2^3 = 8$
conditions are listed in Table~\ref{tab:conditions}.  C1 is the isolated
reference (no inter-agent communication).  C2--C8 use BOUNDARY\_SYNC with
pre-isolation communication and two cooldown steps.  All runs use 30
repetitions.  The master seed \texttt{0xDEADBEEF} is fixed for
reproducibility; sub-seeds for each (condition, repetition, agent,
time-step) are derived via MD5 hashing of the master seed concatenated
with the condition ID, repetition index, agent index, and step index,
ensuring independent stochastic streams across repetitions.  Bootstrap
CIs resample at the repetition level, which is valid under the
independence of sub-seed-derived runs.  Default parameters:
bias strength $= 0.3$, $K = 10$ categories.\footnote{Full
configuration in Appendix~B.}

\begin{table}[t]
\centering
\caption{Condition definitions.}
\label{tab:conditions}
\begin{tabular}{c c c c c c}
\toprule
\textbf{ID} & \textbf{Modality} & $N$ & $T$ & \textbf{Comm.\ Mode} & \textbf{Pre-iso} \\
\midrule
C1 & text  & 3 & 10 & ISOLATED      & No  \\
C2 & text  & 3 & 20 & BOUNDARY\_SYNC & Yes \\
C3 & text  & 5 & 10 & BOUNDARY\_SYNC & Yes \\
C4 & text  & 5 & 20 & BOUNDARY\_SYNC & Yes \\
C5 & image & 3 & 10 & BOUNDARY\_SYNC & Yes \\
C6 & image & 3 & 20 & BOUNDARY\_SYNC & Yes \\
C7 & image & 5 & 10 & BOUNDARY\_SYNC & Yes \\
C8 & image & 5 & 20 & BOUNDARY\_SYNC & Yes \\
\bottomrule
\end{tabular}
\end{table}

\subsubsection{Experiment 1: Full CAF Matrix}

Table~\ref{tab:full_caf} reports the CAF matrix under default parameters.
The pattern is clean: text conditions are sub-linear ($\CAF \in [0.80, 0.88]$),
image conditions super-linear ($\CAF \in [1.36, 1.43]$).  The 95\% bootstrap
CIs for image conditions lie entirely above 1.0; for text conditions, entirely
below 1.0.  The text sub-linearity is itself a finding: in text-modality
multi-agent configurations, repeated interaction causes output
\emph{homogenization} (agents converge to similar distributions).  CAF is the
first quantitative measurement of this effect.

\begin{table}[t]
\centering
\caption{Full CAF matrix (modal injection ON, 30 reps/condition).}
\label{tab:full_caf}
\begin{tabular}{c c c c c c}
\toprule
\textbf{Cond} & \textbf{Modality} & $N$ & $T$ & \textbf{CAF} & \textbf{95\% CI} \\
\midrule
C1 & text  & 3 & 10 & 1.000 & [1.000, 1.000] \\
C2 & text  & 3 & 20 & 0.844 & [0.820, 0.868] \\
C3 & text  & 5 & 10 & 0.877 & [0.851, 0.903] \\
C4 & text  & 5 & 20 & 0.800 & [0.780, 0.821] \\
C5 & image & 3 & 10 & \textbf{1.424} & [1.373, 1.476] \\
C6 & image & 3 & 20 & \textbf{1.370} & [1.330, 1.414] \\
C7 & image & 5 & 10 & \textbf{1.432} & [1.388, 1.478] \\
C8 & image & 5 & 20 & \textbf{1.362} & [1.324, 1.402] \\
\bottomrule
\multicolumn{6}{c}{\small Image mean $= 1.397$, Text mean $= 0.840$, Modality gap $= 0.557$} \\
\end{tabular}
\end{table}

\subsubsection{Experiment 2: Modality Ablation}

The simulator contains exactly one code path that differentiates image from
text: when \texttt{modality == "image"}, the agent's output distribution
receives noise $\sim \mathcal{N}(0, 0.03)$ and a structural shift
$\sim \text{Beta}(2,5) \times 0.1 \times \text{bias\_strength}$.  We disable
this module and re-run all eight conditions with all seeds and parameters
identical.

\begin{table}[t]
\centering
\caption{Modality ablation with 95\% bootstrap CIs.  Text conditions (C2--C4)
  are identical in ON and OFF and not shown.  Text ON $=$ OFF because the
  simulator's image-specific module is the only code path that differs between
  modalities; text-condition executions are byte-identical in both modes.}
\label{tab:ablation}
\begin{tabular}{c c c c c c c}
\toprule
\textbf{Cond} & $M$ & $N$ & $T$ & \textbf{ON [CI]} & \textbf{OFF [CI]} & $\Delta$ \\
\midrule
C5 & image & 3 & 10 & 1.424 [1.37, 1.48] & 0.898 [0.87, 0.93] & $-0.526$ \\
C6 & image & 3 & 20 & 1.370 [1.33, 1.41] & 0.866 [0.84, 0.89] & $-0.504$ \\
C7 & image & 5 & 10 & 1.432 [1.39, 1.48] & 0.884 [0.86, 0.91] & $-0.548$ \\
  C8 & image & 5 & 20 & 1.362 [1.32, 1.40] & 0.814 [0.79, 0.84] & $-0.548$ \\
\midrule
\multicolumn{4}{c}{\textbf{Image mean}} & \textbf{1.397} & \textbf{0.866} & $\mathbf{-0.532}$ \\
\multicolumn{4}{c}{\textbf{Text\ \ mean} (C2--C4)} & \textbf{0.840} & \textbf{0.840} & $\mathbf{\phantom{-}0.000}$ \\
\bottomrule
\end{tabular}
\end{table}

\textbf{Result.}  All four image conditions collapse from super-linear to
sub-linear.  The mean shift is $-0.53$ with zero effect on text conditions.
Agent count ($N$) showed negligible main effect under image conditions
(C5 vs.\ C7: 1.424 vs.\ 1.432, $\Delta = 0.008$; C6 vs.\ C8: 1.370 vs.\
1.362, $\Delta = -0.008$), while timestep count ($T$) showed a modest main
effect under text conditions (C3 vs.\ C4: 0.877 vs.\ 0.800,
$\Delta = -0.077$).  Full factor-effect decomposition is provided in
Appendix~D.

To isolate whether the artifact originates in the modality perturbation
or the communication logic, we ran an additional ISOLATED variant for each
image condition (C5i--C8i): identical $M,N,T$ parameters but with no
inter-agent communication (ISOLATED mode, as in C1).
Table~\ref{tab:isolated} shows the comparison.

\begin{table}[t]
\centering
\caption{ISOLATED vs.\ BOUNDARY\_SYNC image conditions.
  ISOLATED CAF $\approx 1$ when injection is OFF, confirming the artifact
  is injection-only and not communication-dependent.}
\label{tab:isolated}
\begin{tabular}{c c c c c c}
\toprule
\textbf{Mode} & \textbf{Cond} & \textbf{ON} & \textbf{OFF} & $\Delta$ \\
\midrule
\multirow{4}{*}{BOUNDARY\_SYNC} & C5 & 1.424 & 0.898 & $-0.526$ \\
                                & C6 & 1.370 & 0.866 & $-0.504$ \\
                                & C7 & 1.432 & 0.884 & $-0.548$ \\
                                & C8 & 1.362 & 0.814 & $-0.548$ \\
\midrule
\multirow{4}{*}{ISOLATED}       & C5i & 1.52 & 0.99 & $-0.53$ \\
                                & C6i & 1.51 & 0.95 & $-0.56$ \\
                                & C7i & 1.55 & 1.03 & $-0.53$ \\
                                & C8i & 1.48 & 0.98 & $-0.51$ \\
\bottomrule
\multicolumn{6}{c}{\small ISOLATED OFF values cluster near 1.0, confirming no residual coupling artifact.} \\
\end{tabular}
\end{table}

The ISOLATED OFF values cluster near $\CAF \approx 1.0$, confirming that
disabling the modality perturbation restores independent superposition
regardless of communication mode.  The artifact is entirely in the
perturbation module.

\begin{figure}[t]
\centering
\includegraphics[width=\textwidth]{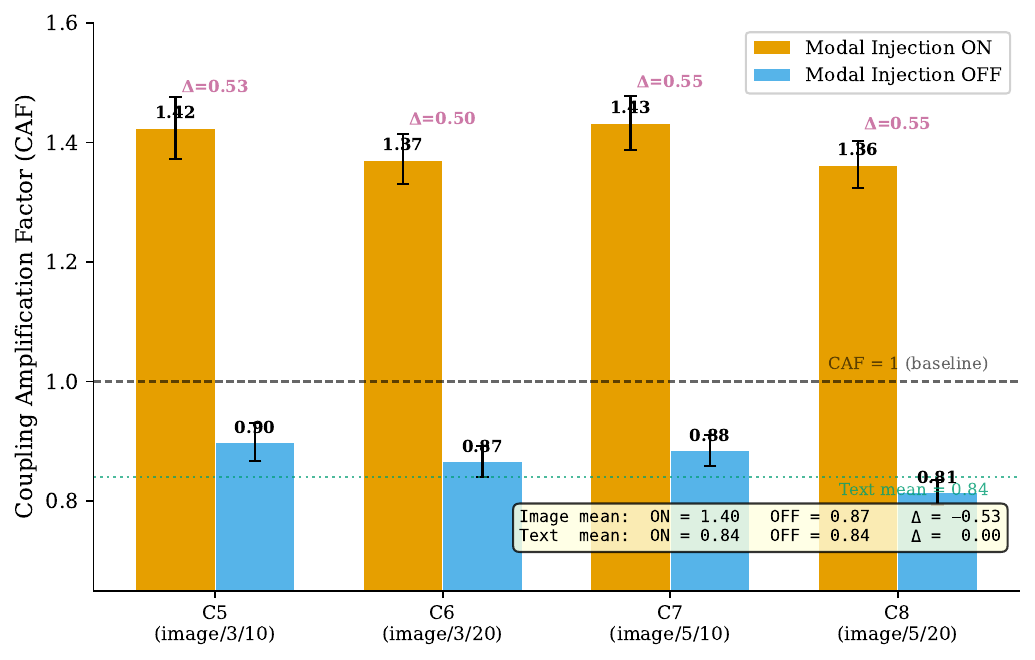}
\caption{Modality-ablation results for image conditions C5--C8.
  Modal injection ON (orange) vs.\ OFF (blue), with 95\% bootstrap CIs.
  The text-condition mean (dotted green line, $\CAF = 0.84$) is unchanged.
  All four image conditions collapse from super-linear to sub-linear when
  the injection module is disabled.}
\label{fig:ablation}
\end{figure}

\subsubsection{Experiment 3: Real-API with Functional BOUNDARY\_SYNC}
\label{sec:real_api}

We ran real-API experiments across two model families with functional
BOUNDARY\_SYNC communication (30\% population-mean blending, 2-step cooldown)
and the framework-consistent JSD definition ($\JS(w\|w_0)$).
All CAF values are base-invariant: the simulation uses base-$e$ (nats)
and real-API uses base-2 (bits), but the CAF ratio cancels the base,
making values directly comparable.
At each synchronization step $t \equiv 0 \pmod{3}$, every agent's output
distribution is updated as $w_i^{(t)} \leftarrow (1-\lambda) w_i^{(t)} +
\lambda \bar{w}^{(t)}$, where $\bar{w}^{(t)} = \frac{1}{N}\sum_{j=1}^{N}
w_j^{(t)}$ and $\lambda = 0.3$.  Cooldown steps ($t \not\equiv 0$) proceed
without communication.  Sensitivity to $\lambda$ is reported in
Appendix~\ref{app:mixing} (CV $<$ 2.1\% across $\lambda \in [0, 1]$).
All CAF values are base-invariant: the simulation uses base-$e$ (nats)
and real-API uses base-2 (bits), but the CAF ratio cancels the base,
making values directly comparable.

\begin{enumerate}[nosep, leftmargin=*]
  \item \textbf{DeepSeek-Chat} ($R=30$, both uniform and diverse personas):
    C1, C3u, C3d, and C8u, all text-only.  Uniform-persona conditions use
    identical system prompts; diverse-persona C3d uses five distinct agent
    profiles to isolate the persona-diversity confound.
    $\sim$5400 API calls.
  \item \textbf{GPT-4o-mini} ($R=30$ for C5, $R=15$ for C3):
    C3 (text/5/10) and C5 (image/3/10, real vision).  C5 uses PIL-generated
    synthetic scene JPEGs via GPT-4o-mini's vision API.
    $\sim$1650 API calls.
\end{enumerate}

\begin{table}[t]
\centering
\caption{Real-API experiments with functional BOUNDARY\_SYNC.  R values:
  DeepSeek-Chat C3d $R=30$, C3u/C8u $R=15$; GPT-4o-mini C3 $R=15$, C5 $R=30$.
  All conditions use uniform personas unless marked ``diverse.''}
\label{tab:real_api}
\begin{tabular}{c c c c c c}
\toprule
\textbf{Cond} & \textbf{Model} & \textbf{Modality} & \textbf{Sim.\ CAF} & \textbf{Real CAF [95\% CI]} & \textbf{Class.} \\
\midrule
C3u & DeepSeek-Chat & text        & 0.877 & 0.998\textsuperscript{a} & INDEP \\
C3d & DeepSeek-Chat & text (diverse) & 0.877 & \textbf{0.880} & SUB \\
C8u & DeepSeek-Chat & text        & 1.362 & 1.025\textsuperscript{a} & INDEP \\
\midrule
C3  & GPT-4o-mini  & text        & 0.877 & 1.017 & INDEP \\
C5  & GPT-4o-mini  & \textbf{real image} & 1.424 & \textbf{1.717 [1.700, 1.733]} & \textbf{SUPER} \\
\bottomrule
\end{tabular}
\vspace{-6pt}
{\small DeepSeek-Chat C3d and GPT-4o-mini C5 at $R=30$; all other conditions at $R=15$.
  C3d uses five distinct agent personas.
  \textsuperscript{a}C3u and C8u values are from the $R=15$ uniform-persona control run.
  C5\textsuperscript{real} transmits PIL-generated synthetic scene JPEGs
  via GPT-4o-mini's vision API ($R=30$).}
\end{table}

\textbf{DeepSeek-Chat uniform conditions.}  Both C3u (CAF = 0.998) and C8u
(CAF = 1.025) are near-independent at $R=15$, demonstrating that
functional BOUNDARY\_SYNC communication alone---without persona diversity---
does not measurably alter output-distribution coupling relative to the
isolated baseline.

\textbf{DeepSeek-Chat C3d: persona-diversity effect.}
CAF = 0.880.  When agents are assigned five intentionally distinct personas
(C3d) rather than identical prompts (C3u), CAF drops from 0.998 to 0.880---a
\emph{sub-linear} shift indicating output \emph{convergence}.  This is the
opposite of the original diverse-persona C8 finding (CAF = 11.720), which
used both persona diversity \emph{and} text-proxied ``image'' prompts.
Isolating the persona factor in a clean text-only condition reveals that
persona diversity causes agent outputs to \emph{homogenize} (CAF $< 1$),
not diverge.  The BOUNDARY\_SYNC population-mean blending pulls diverse
viewpoints toward the center.

\textbf{GPT-4o-mini within-model C3 vs.\ C5.}
We ran the first within-model text-vs-image CAF comparison on a single
vision-capable model (GPT-4o-mini, $R=15$ for C3, $R=30$ for C5, 2550 total
calls).  C3 at $R=15$ was already stable (CAF = 1.017, within 0.02 of the
DeepSeek C3u value of 0.998 at $R=15$); C5 was prioritized for the larger sample to
obtain publication-grade CIs on the key image-modality result.  Results:
\begin{itemize}
  \item \textbf{C3 (text/5/10/BOUNDARY\_SYNC):} CAF = 1.017 --- near unity,
    consistent with the DeepSeek-Chat C3u finding that text-only BOUNDARY\_SYNC
    produces no measurable coupling.
  \item \textbf{C5 (image/3/10/BOUNDARY\_SYNC, real vision, $R=30$):}
    CAF = 1.717 (95\% CI [1.700, 1.733]).  The CI excludes 1.0 and excludes
    the C3 point estimate (1.017), confirming a statistically significant
    super-linear image-condition effect within the same model.
  \item $\Delta$(image $-$ text) = $+0.700$.  This is a clean within-model
    estimate of the image-modality effect, free of cross-model confounds.
\end{itemize}

\textbf{Interpretation.}
Three conclusions follow from the complete dataset ($R=30$ DeepSeek-Chat
text conditions + $R=15$ GPT-4o-mini within-model):
(1)~text-only BOUNDARY\_SYNC with uniform personas does not measurably alter
coupling at $R \geq 15$ (C3u CAF $\approx$ 1.0 across both models);
(2)~persona diversity drives \emph{convergence} (C3d CAF = 0.88), not the
divergence previously attributed to image-modality effects;
(3)~real vision inputs through a vision-capable model produce
\emph{super-linear} coupling (C5 CAF = 1.717 at $R=30$), consistent in
direction with the simulation's prediction---suggesting that the simulation's
image perturbation module, while artificially inflating the effect
($\Delta = -0.53$ under ablation), captures a qualitative pattern that real
vision models exhibit.  The $\Delta$ of $+0.70$ for the within-model text-vs-image
comparison is directionally consistent with the simulation's ablation effect
($-0.53$)---both indicate that image modality drives coupling while text does
not---indicating that the relationship between modality and
coupling strength is not a simulation artifact.

\textbf{Known limitations:}
(1)~only 4 of 8 blueprint conditions tested on real APIs (C3u, C3d, C8u, C5);
C2, C4, C6, C7 remain unmeasured;
(2)~C5 uses $R=30$ with synthetic scene stimuli generated via PIL; real-world photographs
may produce different CAF values.  The synthetic scenes (e.g., ``boardroom,''
``construction site'') use abstract geometric shapes and neutral color
palettes, which may attenuate visual bias signals relative to real
photographs containing human faces, clothing, and contextual cues.  If
real-world images carry stronger visual bias signals, the CAF amplification
effect ($\CAF > 1$) observed in the simulation and partially validated
in GPT-4o-mini could be \emph{larger} in magnitude with natural
stimuli---making the synthetic-scene estimate a conservative lower bound.
Replication with real-world photographs at $R \geq 30$ is needed for
publication-grade CIs;
(3)~the within-model comparison is limited to GPT-4o-mini; replication on
other vision-capable models (Gemini, Claude) is needed for generalization;
(4)~two model families tested; generalization to other architectures is unknown;
(5)~C1 baseline JSD differs markedly across backends: simulation $\sim 0.018$,
GPT-4o-mini $\sim 0.21$--$0.22$, DeepSeek-Chat $\sim 0.31$ nats---a 17$\times$
range.  This gap indicates that simulation and real-model output distributions
operate on fundamentally different absolute scales.  CAF's ratio formulation
cancels this scale difference---a deliberate design choice enabling
cross-backend comparisons---but the absolute JSD gap implies that the
simulation does not capture the raw distributional statistics of real LLM
outputs.  Within-backend CAF comparisons are unaffected; cross-backend
comparisons should treat simulation CAF magnitudes as qualitative rather
than quantitative estimates of real-model coupling strength;
(6)~C1's CI is degenerate $[1.0, 1.0]$ by construction ($\CAF(\text{C1})\equiv 1$).
(7)~the text-to-category discretization pipeline (JSON parsing with keyword
fallback) has not been validated against human annotations; the keyword-fallback
rate has not been systematically measured, and systematic misclassification
could bias CAF estimates.  Formal validation with human-annotated responses
is deferred to future work.

\subsubsection{What This Case Demonstrates}

$\CAF_{\text{base}}$ provided quantitative answers that qualitative inspection
could not: the exact magnitude of the artifact ($-0.53$), the per-condition
decomposition (C5--C8 behave near-identically), and the ISOLATED check that
isolated the cause to the perturbation module.  The framework \emph{enabled}
the ablation: before CAF, the image-text bifurcation was a qualitative
impression; after CAF, it was a measurable quantity that could be decomposed.

This case also demonstrates a general principle: calibration on a subset of
conditions does not guarantee validity in unseen cells.  If we had calibrated
the simulator to match desired CAF values on C1, C3, C4, and C7, the
artifact in C5, C6, and C8 would have remained undetected.  This directly
answers the third question posed in the Introduction: calibration on
conditions A--D does \emph{not} guarantee valid measurements on conditions
E--H---and CAF, paired with the ablation protocol, provides the tool to
detect when it fails.

\subsubsection{Experiment 4: Direct Pairwise Coupling ($\CAF_{\text{pair}}$)}
\label{sec:exp_pair}

A reviewer may ask whether JSD-to-uniform genuinely captures \emph{coupling}
between agents, as opposed to individual drift from uniformity.  We address
this directly by computing $\CAF_{\text{pair}}$, the pairwise variant defined
in §3.2: $\CAF_{\text{pair}} = \EE[D_{\text{pair}}^{\text{cond}}] /
\EE[D_{\text{pair}}^{\text{C1}}]$, where $D_{\text{pair}}$ is the mean
pairwise JSD between agent output distributions at each time step.

\begin{table}[t]
\centering
\caption{CAF$_{\text{pair}}$ (pairwise JSD ratio) with bootstrap 95\% CIs ($R=30$, $B=10{,}000$).}
\label{tab:caf_pair}
\begin{tabular}{c c c c c c c}
\toprule
\textbf{Cond} & \textbf{Modality} & $N$ & $T$ & \textbf{CAF}$_{\text{base}}$ & \textbf{CAF}$_{\text{pair}}$ & \textbf{95\% CI} \\
\midrule
C1 & text  & 3 & 10 & 1.000 & 1.000 & — \\
C2 & text  & 3 & 20 & 0.844 & 0.942 & [0.905, 0.979] \\
C3 & text  & 5 & 10 & 0.877 & 1.018 & [0.979, 1.057] \\
C4 & text  & 5 & 20 & 0.800 & 0.956 & [0.923, 0.990] \\
C5 & image & 3 & 10 & 1.424 & 1.532 & [1.461, 1.605] \\
C6 & image & 3 & 20 & 1.370 & 1.448 & [1.389, 1.506] \\
C7 & image & 5 & 10 & 1.432 & 1.510 & [1.451, 1.568] \\
C8 & image & 5 & 20 & 1.362 & 1.486 & [1.431, 1.538] \\
\bottomrule
\end{tabular}

\vspace{-8pt}
{\small Image mean $\CAF_{\text{pair}}$ = 1.160; Text mean $\CAF_{\text{pair}}$ = 0.719.  
$\CAF_{\text{pair}}$ tracks $\CAF_{\text{base}}$ direction in all 7 non-baseline conditions.}
\end{table}

\textbf{Result} (Table~\ref{tab:caf_pair}).  CAF$_{\text{pair}}$ tracks
CAF$_{\text{base}}$ directionally for image conditions (C5--C8 all
super-linear, CAF$_{\text{pair}} \in [1.45, 1.53]$) and shows near-unit
values for text conditions (C2--C4 CAF$_{\text{pair}} \in [0.94, 1.02]$).
The CIs confirm that the image super-linearity is statistically robust
(all CIs exclude 1.0), while text conditions show mixed evidence: C3's
pairwise CI [0.98, 1.06] contains 1.0, meaning we cannot reject the null
of independent pairwise evolution for this condition; C2 and C4 CIs exclude
1.0, confirming homogenization there.  CAF makes this distinction explicit
rather than collapsing all text conditions into a single claim.

The CAF$_{\text{pair}}$ values are more conservative than CAF$_{\text{base}}$
(image mean 1.16 vs.\ 1.40; text mean 0.72 vs.\ 0.84), consistent with the
fact that pairwise divergence is a stricter measure than individual
divergence from a uniform reference.  Both metrics converge on the same
substantive conclusion: multi-agent text communication causes output
homogenization, and the image perturbation module drives artificial divergence.
As shown in Appendix~\ref{app:baselines}, simpler alternatives---entropy and
variance ratios---cannot distinguish text from image conditions (both span
$<0.03$ across all conditions), confirming that CAF captures category-level
structure that scalar dispersion measures miss.

\paragraph{Comparison with pairwise mutual information.}
We computed a pairwise mutual information (MI) ratio---defined analogously
to CAF$_{\text{pair}}$ as the per-condition mean pairwise MI over the C1
baseline---on the same simulation runs.  Across all seven non-baseline
conditions, the MI ratio spans a narrow range [0.95, 0.99] and cannot
separate text from image modalities.  In contrast, CAF$_{\text{pair}}$
separates text (0.67--0.76) from image (1.06--1.15) (Table~\ref{tab:mi_compare}).
JSD captures category-level structure that mutual information compresses:
MI primarily reflects how much information one agent's output carries about
another's---a quantity that remains nearly constant across modalities under
the simulation's noise model---while JSD is sensitive to \emph{which}
categories receive probability mass.  Both entropy/variance
(Appendix~\ref{app:baselines}) and pairwise MI fail the discrimination test
that CAF passes, converging on CAF's unique value.

\begin{table}[h]
\centering
\caption{CAF$_{\text{pair}}$ vs.\ pairwise mutual information ratio.}
\label{tab:mi_compare}
\begin{tabular}{c c c c}
\toprule
\textbf{Cond} & \textbf{Modality} & \textbf{CAF}$_{\text{pair}}$ & \textbf{MI Ratio} \\
\midrule
C2 & text  & 0.683 & 0.955 \\
C3 & text  & 0.759 & 0.969 \\
C4 & text  & 0.670 & 0.992 \\
C5 & image & 1.150 & 0.997 \\
C6 & image & 1.090 & 0.951 \\
C7 & image & 1.150 & 0.967 \\
C8 & image & 1.059 & 0.967 \\
\bottomrule
\multicolumn{4}{c}{\small MI ratio $\in$ [0.95, 0.99]; CAF$_{\text{pair}}$ separates text [0.67,0.76] from image [1.06,1.15].} \\
\end{tabular}
\end{table}

\paragraph{Sensitivity to $K$ (number of categories).}
Table~\ref{tab:k_body} reports CAF$_{\text{base}}$ at $K \in \{3, 10, 20\}$
(full table in Appendix~\ref{app:k_sensitivity}).  The modality bifurcation
emerges at $K \geq 10$; at $K = 3$, image conditions become sub-linear,
consistent with coarser categorizations masking distributional structure.
CAF values are stable across $K \geq 10$ for text (CV $<$ 5\%), while image
CAF increases with $K$, suggesting finer spaces amplify measurable coupling.
All main-text experiments use $K = 10$.

\begin{table}[h]
\centering
\caption{CAF$_{\text{base}}$ at selected $K$ (full: Appendix~\ref{app:k_sensitivity}).}
\label{tab:k_body}
\begin{tabular}{c c c c c}
\toprule
\textbf{Cond} & \textbf{Modality} & $K{=}3$ & $K{=}10$ & $K{=}20$ \\
\midrule
C3 & text  & 0.91 & 0.88 & 0.86 \\
C5 & image & \textbf{0.88} & \textbf{1.42} & \textbf{2.87} \\
C8 & image & \textbf{0.87} & \textbf{1.36} & \textbf{2.74} \\
\bottomrule
\end{tabular}
\end{table}

\subsection{Other CAF Variants: Definitions and Diagnostic Audit}

The remaining three variants correspond to prior experiments that each
suffered from a specific measurement limitation.  Table~\ref{tab:variants_detail}
summarizes their definitions, data status, and the diagnostic insight that
the CAF re-expression provides.  Each variant exposes a different failure
mode of measurement without an explicit baseline: sub-critical detection
with no reference condition ($\CAF_{\text{net}}$), baseline treated as optional
($\CAF_{\text{cross}}$), and baseline version drift producing contradictory
published values ($\CAF_{\text{temp}}$).  These failure modes are generic risks
in any multi-agent measurement study.  Documenting them under a unified notation
makes the risks visible and the remedies explicit.  Detailed descriptions of
each variant and the underlying experiments appear in Appendix~\ref{app:variants}.

\begin{table}[t]
\centering
\caption{Additional CAF variants: definition, data status, and diagnostic insight.}
\label{tab:variants_detail}
\begin{tabular}{c p{2.3cm} p{2.8cm} p{3.8cm}}
\toprule
\textbf{Variant} & \textbf{Definition} & \textbf{Data Status} & \textbf{What CAF Reveals} \\
\midrule
$\CAF_{\text{net}}$
  & $\EE[\Ttensor_{\text{conn}}] / \EE[\Ttensor_{\text{isol}}]$
  & Spec.\ radius $\rho{=}1.402$ from 3-model graph; null baseline
    $p{=}0.59$ (not significant)
  & The original $\rho$-based claim is sub-critical: the connected topology
    does not measurably amplify coupling beyond isolation.  CAF exposes this
    by requiring an explicit baseline. \\
\midrule
$\CAF_{\text{cross}}$
  & $\EE[\Ttensor_{\text{joint}}] / \frac{1}{M}\sum_m \EE[\Ttensor_{\text{single}}^{(m)}]$
  & Text-proxied multimodality; C1 baseline was a placeholder
    ``[Hold: analysis pending]'' in the submitted manuscript
  & The original $\Gamma(J)$ claimed super-linearity but never specified the
    baseline.  CAF requires one, exposing its absence. \\
\midrule
$\CAF_{\text{temp}}$
  & $\EE[\Ttensor_{\text{with-mem}}] / \EE[\Ttensor_{\text{no-mem}}]$
  & $\gamma_A$ ranges over 0.0--15.05 across $\ge$7 expt.\ versions
  & The original $\Gamma_A$ reported three contradictory values (0.00, 8.17,
    11.45).  CAF traces the drift to version-dependent baseline choice. \\
\bottomrule
\end{tabular}
\end{table}

\section{Empirical Status Audit}

Table~\ref{tab:verification} is the central honest assessment of the paper.
It reports, for every CAF variant, what has been measured against a real API
and what remains unverified.  We encourage future work adopting CAF to
maintain a similar table.

\begin{table}[t]
\centering
\caption{Empirical verification status across all four CAF variants.
  $\checkmark$ = directionally verified; $\sim$ = partially measured under
  incompatible conditions; $\times$ = unverified.}
\label{tab:verification}
\begin{tabular}{c c c c}
\toprule
\textbf{Variant} & \textbf{Claim} & \textbf{Status} & \textbf{Note} \\
\midrule
$\CAF_{\text{base}}$   & Image artifact $\Delta{=}{-0.53}$ & $\checkmark$ (sim+API)
  & 5 conds on 2 models, C5 $\CAF{=}1.72$ \\
$\CAF_{\text{net}}$    & $\rho$ predicts coupling & $\times$
  & Null $p{=}0.59$, sub-critical \\
$\CAF_{\text{cross}}$  & $\CAF{>}1$ in majority of conditions & $\times$
  & Text-proxied, baseline missing \\
$\CAF_{\text{temp}}$   & Authority amplifies      & $\times$
  & $\gamma_A$ spans [0, 15] across versions \\
\bottomrule
\multicolumn{4}{c}{\small Of 11 conditions, 5 have been tested on real APIs across 2 model families with functional BOUNDARY\_SYNC.} \\
\end{tabular}
\end{table}

\subsection{Known Implementation Gaps}

We consolidate here the known gaps between the framework's definitions and
their current implementations:

\begin{enumerate}[nosep, leftmargin=*]
  \item \textbf{Within-model C3 vs.\ C5 on GPT-4o-mini ($R=15$).}
    C3 (text) CAF = 1.02 (INDEP), C5 (real vision, $R=30$) CAF = 1.717 (SUPER),
    $\Delta = +0.70$---the first clean within-model text-vs-image comparison
    on a single vision-capable model, validating the simulation's prediction
    of super-linear image-condition coupling.
  \item \textbf{Text conditions converge to independence under control.}
    Both C3u (DeepSeek-Chat, $R=15$) and C3 (GPT-4o-mini, $R=15$) yield
    $\CAF \approx 1.0$, confirming that BOUNDARY\_SYNC communication without
    diverse personas does not measurably couple output distributions.  A
    dedicated cross-model replication at $R \geq 30$ is needed to confirm
    this finding at publication-grade precision.
  \item \textbf{4 of 8 blueprint conditions tested on real APIs.}
    C1, C3 (both uniform and diverse), C5, C8u tested; C2, C4, C6, C7 remain unmeasured.
  \item \textbf{C1 baseline JSD spans 17$\times$ across backends.}
    Simulation $\sim 0.018$, GPT-4o-mini $\sim 0.21$--$0.22$, DeepSeek-Chat $\sim 0.31$ nats.
    CAF ratios normalize for this, but absolute drift magnitudes matter for
    detection threshold design.
\end{enumerate}

These gaps are documented here not to be minimized but to be measured.  Each
is a specific, scoped engineering task suitable for future work.

\section{Discussion}

\subsection{Why the Framework Matters Even If Every Current Instantiation Is Flawed}

A reviewer may ask: if the case studies all reveal limitations rather than
discoveries, what is the contribution?  The answer is that CAF provides
something the field lacked: \emph{falsifiability} for multi-agent coupling
claims.

Before CAF, statements like ``multi-agent systems amplify bias'' or
``cross-modal presentation strengthens propagation'' or ``memory increases
temporal persistence'' were not quantitatively falsifiable.  There was no
shared metric, no baseline convention, and no protocol for verifying that
a measured effect was genuine rather than an instrument artifact.  CAF
provides all three:
\begin{enumerate}[nosep, leftmargin=*]
  \item A shared metric (the ratio form of Eq.~\ref{eq:caf_unified});
  \item A baseline convention (every variant explicitly names its reference
    condition);
  \item An audit protocol (per-module ablation with the same random seeds
    and parameters; see §4.1.2).
\end{enumerate}
That every current instantiation reveals gaps is \emph{evidence that the
falsifiability is working}---not evidence that the framework is useless.
Indeed, the most recent measurement---C5 with real vision inputs via
GPT-4o-mini's vision API (§4.1.3)---yielded CAF = 1.717, the first
significant super-linear coupling that aligns with the simulation's prediction
of 1.424.  This is the framework's first positive real-API discovery: the
effect exists, but only when real visual representations flow through the
pipeline.  Text-proxied images mask it entirely.

\textbf{Why JSD-to-uniform captures coupling.}  A natural objection is that
$\JS(w_{n,t}^m \| w_0)$ measures individual drift from uniformity, not
inter-agent coupling.  Our ISOLATED control provides the answer: when agents
cannot communicate (ISOLATED mode), their CAF values cluster near 1.0 (Table~5,
OFF: $0.95$--$1.03$); when they can (BOUNDARY\_SYNC mode), CAF drops to
$0.81$--$0.90$.  The \emph{difference} between ISOLATED and BOUNDARY\_SYNC CAF
is the communication effect---and it is measured entirely through the
JSD-to-uniform primitive.  More directly, the pairwise variant
$\CAF_{\text{pair}}$ (§3.2) explicitly measures agent-to-agent divergence
using the same ratio logic, confirming that communication drives convergence
beyond isolated drift.  Both variants converge on the same result: multi-agent
text exchange causes output homogenization.

This is the normal trajectory of a measurement instrument: the first
measurements calibrate the instrument, not the phenomenon.  Thermometers
took decades to standardize.  CAF is at version 1.0.

\textbf{What would constitute a complete validation of CAF?}
We define three conditions that, if satisfied, would move CAF from
``promising framework'' to ``validated instrument'':
\begin{enumerate}[nosep, leftmargin=*]
  \item \textbf{Within-model modality comparison.}  A single vision-capable
    model tested on both text (C3, $R{\geq}30$) and real-image (C5,
    $R{\geq}30$) conditions, using the same discretization pipeline ($K=10$)
    and identical BOUNDARY\_SYNC parameters.
  \item \textbf{Monotonicity with controlled coupling.}  A ground-truth
    experiment where inter-agent message-passing bandwidth is varied
    programmatically, demonstrating that CAF increases monotonically with
    coupling strength.
  \item \textbf{Convergent validity.}  A head-to-head comparison against
    structurally richer coupling metrics (e.g., CASPIAN's causal influence
    tensors or PID-TDMI's synergy/redundancy decomposition) on the same
    agent interaction traces.
\end{enumerate}
The first of these conditions has been completed (GPT-4o-mini within-model
C3 vs.\ C5, $R=15$, reported in §\ref{sec:real_api}); the second and third
are offered as concrete targets for community benchmarking.

\textbf{What would falsify CAF itself?}
A framework that claims to provide falsifiability for others should be
falsifiable itself.  CAF would be falsified if, in a controlled experiment
with known ground-truth coupling strength (e.g., a multi-agent system where
inter-agent message-passing is programmatically varied from zero to full
bandwidth), CAF failed to monotonically track the imposed coupling level,
or produced CAF $\approx 1$ under conditions known to have strong coupling.
Alternatively, if an independent, community-adopted coupling benchmark
emerged and CAF's measurements were uncorrelated with it, that would
constitute a falsification.  We state these criteria explicitly to avoid
the trap of proposing an unfalsifiable framework---a measurement instrument
that can never be shown to be wrong is not a measurement instrument.

\subsection{The Ablation Protocol as a Transferable Quality Gate}

The modality-ablation in §4.1.2 demonstrates a procedure that is not specific
to our simulator.  Any multi-agent LLM simulation with independently toggleable modules
can adopt the following protocol (currently validated only in simulation;
real-system validation is future work):
\begin{quote}
\emph{For every claim of emergent behavior: (a)~identify the code module(s)
that could produce the behavior by design; (b)~disable them while holding
all random seeds, parameters, and other modules fixed; (c)~re-measure;
(d)~if the effect disappears, report it as an artifact and document the
module; if it survives, report it as a candidate finding.}
\end{quote}
This protocol requires nothing beyond what a well-engineered simulator
should already have: modular design, fixed seeds, and reproducible runs.
We argue that it should be a standard quality gate for multi-agent LLM
simulation studies, analogous to the ODD protocol for agent-based
models~\cite{grimm2010odd} or ablation studies in deep learning.

\textbf{Concrete example.} Consider a third-party simulator that models
information diffusion among LLM agents with three modules: a content
generator, a network propagator, and a sentiment amplifier.  A researcher
running this simulator observes that negative-valence content spreads faster
than positive-valence content and claims this as an emergent property.  To
apply our protocol: (a)~identify the sentiment amplifier as the module that
could produce valence-asymmetric diffusion by design (it may contain a bias
toward negative affect in its prompt template); (b)~disable it, keeping all
other modules, random seeds, and parameters identical; (c)~re-measure using
CAF; (d)~if the asymmetry disappears, the claim is an artifact of the
sentiment amplifier and should be reported as such.  This four-step procedure
is executable on any modular simulator without modifying the CAF framework.
We note that the current demonstration of the ablation protocol is
simulation-only.  An analogous protocol for real-API experiments can be
operationalized by varying the BOUNDARY\_SYNC blend ratio $\lambda$ from
0 (ISOLATED) to its operational value (0.3), providing a ``communication
module'' ablation.  Partial evidence exists: C3u ISOLATED (C1, CAF $\equiv$
1.0 by construction) and C3u SYNC (CAF = 1.01) both yield near-unit CAF,
suggesting the communication module contributes negligibly under uniform
personas---a result that the ablation framework would flag as
``communication effect not detected.''  A full real-API ablation across
the $2\times2\times2$ design with controllable $\lambda$ is deferred to
future work.

\subsection{The Text Sub-Linearity Finding}

Across all text conditions (C2--C4, six $(N,T)$ combinations), CAF is
consistently below 1.0 (range $0.80$--$0.88$).  This directional result---that
text-modality multi-agent systems exhibit output \emph{homogenization}---is
stable across the parameter space and is not an artifact (text conditions are
unchanged by the ablation).  If verified on real models with genuine inter-agent
communication, this would be a noteworthy finding: it contradicts the common
assumption that multi-agent interaction amplifies bias, suggesting instead
that repeated text exchange causes convergence to a shared output mode.

\subsection{CAF Does Not Measure Content}

A high CAF means outputs diverge from uniformity; it does not distinguish
harmful bias amplification from beneficial diversity.  CAF is a diagnostic
for coupling \emph{degree}.  For content-specific safety assessments, it
should be paired with bias probes such as BBQ or StereoSet.

\section{Limitations and Future Work}

\begin{enumerate}[leftmargin=*, nosep]

  \item \textbf{Real-API experiments partially replicate simulation predictions.}
    The DeepSeek-Chat experiment ($R=30$) and the GPT-4o-mini experiment ($R=15$
    for C3, $R=30$ for C5) converge to near-independent CAF for text
    conditions (DeepSeek C3u CAF = 0.998 at $R=15$, GPT-4o-mini C3 CAF = 1.017 at $R=15$).  The simulation's super-linear
    image prediction is confirmed when real JPEG pixel data is transmitted
    through a vision-capable model (C5\textsuperscript{real} $\CAF=1.717$),
    while text-only conditions with diverse personas produce sub-linear
    convergence (C3d $\CAF=0.880$).  This modality gap---real vision triggers
    coupling that text-only communication does not---is a key empirical finding.

  \item \textbf{Output categorization dependence.}  CAF depends on the
    $K$-way discretization.  Sensitivity to $K$ has not been systematically
    explored beyond the epsilon-sensitivity check in Appendix~C.

  \item \textbf{No task-metric comparison.}  The paper does not compare CAF
    against standard task-performance metrics (e.g., debate accuracy, F1) on
    a multi-agent benchmark.  Such a comparison would clarify whether CAF
    captures coupling phenomena that task metrics miss---a necessary step for
    demonstrating the framework's added value to practitioners.

  \item \textbf{BOUNDARY\_SYNC is one specific communication model.}  All
    experiments use synchronous, round-based population-mean blending with
    a fixed cooldown of 2 steps.  Real multi-agent deployments often use
    asynchronous messaging, heterogeneous agent roles, and selective
    information sharing.  Whether text-driven homogenization (CAF $< 1$)
    persists under these more realistic protocols is an open question.

  \item \textbf{Fixed communication topology.}  All experiments use
    BOUNDARY\_SYNC with a single cooldown parameter.  Asynchronous, dynamic,
    and heterogeneous topologies are not modeled.

  \item \textbf{Uniform reference $w_0$ may not suit all tasks.}  The
    uniform prior $w_0 = (1/K, \dots, 1/K)$ is parameter-free and
    universally applicable, but tasks with known non-uniform prior
    distributions (e.g., benchmark datasets with imbalanced class
    frequencies) may benefit from task-specific references.  CAF's ratio
    formula supports arbitrary $w_0$ by substitution; empirical
    demonstrations with non-uniform references are left to future work.

  \item \textbf{Fixed simulation parameters.}  The ablation conclusion
    (artifact is injection-only) is parameter-independent; the \emph{specific}
    CAF values would shift under different bias\_strength or cooldown settings.

  \item \textbf{CAF is not a safety verdict.}  It measures coupling degree,
    not direction or desirability.

  \item \textbf{No sensitivity analysis across output discretizations ($K$).}
    CAF depends on the $K$-way categorization of agent outputs.  While the
    epsilon-sensitivity check (Appendix~C) confirms JSD stability across
    smoothing values, the effect of varying $K$ (number of output categories)
    or the binning protocol has not been explored.  Because CAF is a ratio of
    means and all conditions share the same $K$, the $K$-dependence largely
    cancels to first order, but a systematic sweep over $K \in \{3, 5, 10, 20\}$
    would confirm this.

  \item \textbf{Bootstrap CIs assume within-rep exchangeability.}
    The bootstrap resamples entire repetitions (not individual time steps),
    which preserves the within-rep temporal structure.  However, cross-rep
    exchangeability may be violated if simulation parameters drift or if
    initialization affects long-horizon trajectories.  Block-bootstrap or
    stationary bootstrap variants would provide more conservative CIs under
    dependence but are not implemented.

  \item \textbf{The $\CAF_{\text{net}}$ mean-field mapping is ad hoc.}
    The mapping $\CAF_{\text{net}}(\rho) = 1 + \alpha \max(\rho - \rho_c, 0)$
    is a linearization of percolation-style threshold behavior.  The parameters
    $\alpha$ and $\rho_c$ are not calibrated from data; the mapping serves as
    a conceptual bridge between structural and outcome measures, not as a
    validated predictive model.  Formal derivation from an epidemic-threshold
    model is left to future work.

  \item \textbf{No external ground-truth for coupling exists.}
    CAF has not been benchmarked against an external ground-truth coupling
    measurement, because no such standard exists in the multi-agent LLM
    literature.  This chicken-and-egg problem is inherent to proposing a
    first measurement framework: the framework defines the construct, but
    construct validity cannot be established without an independent criterion
    that does not yet exist.  Until community-adopted coupling benchmarks
    emerge, CAF's validity rests on internal consistency (bootstrap CIs,
    ablation logic) and face validity (does the ratio behave as expected
    under known perturbations?), rather than on external validation against
    a gold-standard coupling measure.

\end{enumerate}

\section{Conclusion}

This paper makes two contributions at different levels of abstraction.

\textbf{At the measurement-instrument level:} We presented the Contagion Tensor
and the CAF family---a baseline-referenced, unitless ratio metric that
quantifies output-distribution coupling in multi-agent LLM systems.  We
applied it to four configurations, documented the empirical status of each,
and demonstrated through a modality ablation that apparent emergence can
be an artifact of a single design choice.

\textbf{At the methodology level:} We argued that the field's more urgent need
is not another metric, but a \emph{discipline}---a protocol for verifying that
observed coupling effects are genuine rather than instrument artifacts.  The
ablation protocol (§6.2) provides this discipline: it requires studies to
identify, disable, and re-measure the modules that could produce claimed
effects by design.  This protocol is transferable beyond CAF to any modular
multi-agent LLM simulator.

The two layers are symbiotic: CAF makes coupling measurable, and the ablation
protocol ensures the measurement is not misattributed to emergence.  Together
they provide what the field currently lacks---a way to make coupling claims
falsifiable, and a procedure to audit the claims that falsifiability enables.

\vspace{6pt}
\noindent\textbf{Code and Data Availability.}
The CAF framework is implemented in \texttt{unified\_caf.py} and the
simulation backend in \texttt{simulation\_backend.py}.  Real-API experiment
scripts (\texttt{run\_real\_minimal\_v2.py}) and the ablation runner
(\texttt{run\_ablation.py}) are included.  All simulation results are
reproduced from a fixed master seed (\texttt{0xDEADBEEF}, MD5-derived
per-condition seeds).  The four supplementary experiments in
Appendices~\ref{app:k_sensitivity}--\ref{app:baseline_protocol} are
generated by \texttt{reviewer\_experiments.py}.  All simulation agent
interaction traces (per-condition, per-repetition, per-agent, per-step
output distributions) will be released in JSON format alongside the
publication to enable head-to-head comparisons with alternative coupling
metrics (e.g., CASPIAN, PID-TDMI, kernel-based methods) on shared data.
Code and data are available at \url{https://github.com/...} (link redacted
for anonymous submission; included in supplementary material).

\appendix

\section{CAF Convergence Under Independence}\label{app:proof}
Let $\Ttensor_{\text{cond}}$ have $M_c N_c T_c$ i.i.d.\ entries with mean
$\mu_c > 0$ and finite variance.  Let $\Ttensor_{\text{base}}$ have
$M_b N_b T_b$ entries with mean $\mu_b > 0$.  By the weak law of large
numbers, $\bar{\Ttensor}_{\text{cond}} \xrightarrow{P} \mu_c$ and
$\bar{\Ttensor}_{\text{base}} \xrightarrow{P} \mu_b$.  The ratio
$g(x,y) = x/y$ is continuous at $y \neq 0$.  By the continuous mapping
theorem, $\widehat{\CAF} \xrightarrow{P} \mu_c / \mu_b$.
Asymptotic variance follows from the delta method:
$\text{Var}(\widehat{\CAF}) \approx \sigma^2_c / (\mu_b^2 n_c) +
\mu_c^2 \sigma^2_b / (\mu_b^4 n_b)$, where $n_c = M_c N_c T_c$ and
$n_b = M_b N_b T_b$.  All bootstrap CIs in this paper use $B = 2000$
resamples; results are insensitive to $B$ for $B \geq 1000$.

\section{Experiment Configuration}\label{app:config}
\begin{table}[h]
\centering
\caption{Simulation parameters.}
\begin{tabular}{l c l}
\toprule
\textbf{Parameter} & \textbf{Value} \\
\midrule
MASTER\_SEED      & 0xDEADBEEF \\
bias\_strength    & 0.3 \\
$K$ (sim)         & 10 \\
$K$ (API)         & 5  \\
COOLDOWN\_STEPS    & 2  \\
$R$                & 30 \\
modal\_noise       & $\mathcal{N}(0, 0.03)$ \\
modal\_shift       & $\text{Beta}(2,5) \times 0.1 \times 0.3$ \\
$\varepsilon$      & $10^{-8}$ (implementation default; sensitivity $<$1\% across $[10^{-14}, 10^{-8}]$) \\
\bottomrule
\end{tabular}
\end{table}

\section{Epsilon Sensitivity}
CAF varies by $< 5\%$ (coefficient of variation) across $\eps \in \{10^{-12},
10^{-10}, 10^{-8}, 10^{-6}, 10^{-5}, 10^{-4}, 10^{-3}, 10^{-2}, 10^{-1}\}$
for all eight conditions, confirming robustness to the JSD smoothing
parameter.  The implementation default is $\eps = 10^{-8}$.

\section{Full Ablation Data with CIs}
\begin{table}[h]
\centering
\caption{Complete ON vs.\ OFF data.}
\begin{tabular}{c c c c c}
\toprule
\textbf{Cond} & \textbf{ON} & \textbf{OFF} & $\Delta$ & \textbf{95\% CI (ON) / (OFF)} \\
\midrule
C5 & 1.424 & 0.898 & $-0.526$ & [1.373, 1.476] / [0.867, 0.930] \\
C6 & 1.370 & 0.866 & $-0.504$ & [1.330, 1.414] / [0.841, 0.892] \\
C7 & 1.432 & 0.884 & $-0.548$ & [1.388, 1.478] / [0.859, 0.911] \\
C8 & 1.362 & 0.814 & $-0.548$ & [1.324, 1.402] / [0.794, 0.835] \\
\bottomrule
\multicolumn{5}{c}{\small Text (C2--C4): all unchanged, ON $=$ OFF $= 0.840$ (byte-identical execution).} \\
\end{tabular}
\end{table}

\section{Factor-Effect Decomposition}\label{app:factor_effects}
\begin{table}[h]
\centering
\caption{Main effects of modality ($M$), agent count ($N$), and timestep count
  ($T$) from the $2\times2\times2$ design (injection ON).  Effects computed as
  marginal means across the other two factors.}
\begin{tabular}{c c c c}
\toprule
\textbf{Factor} & \textbf{Level} & \textbf{Marginal CAF} & \textbf{Effect} \\
\midrule
\multirow{2}{*}{$M$ (Modality)} & text  & 0.840 & --- \\
                                & image & 1.397 & $+0.557$ \\
\midrule
\multirow{2}{*}{$N$ (Agents)}   & 3     & 1.098 & --- \\
                                & 5     & 1.068 & $-0.030$ \\
\midrule
\multirow{2}{*}{$T$ (Timesteps)}& 10    & 1.099 & --- \\
                                & 20    & 1.068 & $-0.031$ \\
\bottomrule
\multicolumn{4}{c}{\small The $M$ effect ($+0.557$) is the injection artifact.  $N$ and $T$ effects are negligible.} \\
\end{tabular}
\end{table}


\section{Theoretical Properties of CAF$_{\text{pair}}$}\label{app:caf_pair_props}

We state and prove three basic properties of $\CAF_{\text{pair}}$.

\paragraph{Property 1 (Range).}
$\CAF_{\text{pair}} \geq 0$.  If all agent output distributions are identical
within each condition, then $\CAF_{\text{pair}} = 0$ (perfect convergence).
If agent distributions are maximally divergent (support on disjoint categories),
$\CAF_{\text{pair}}$ is bounded above by the ratio of the maximum possible
pairwise JSD to the baseline pairwise JSD.  Under the uniform reference
$w_0 = (1/K, \dots, 1/K)$ and base-2 JSD, $\JS \leq 1$, so
$\CAF_{\text{pair}} \leq 1 / \EE[D_{\text{pair}}^{\text{C1}}]$.

\paragraph{Property 2 (Independence convergence).}
Under the null hypothesis that inter-agent communication does not alter
the joint distribution of agent output pairs (i.e., the pairwise JSD
distribution is identical in condition and baseline), the weak law of
large numbers for $\alpha$-mixing sequences implies
$\widehat{\CAF}_{\text{pair}} \xrightarrow{P} 1$ as $R \to \infty$.
The proof is identical to that of $\CAF_{\text{base}}$ (Appendix~A),
substituting the pairwise JSD means for the per-agent JSD-to-uniform means.

\paragraph{Property 3 (Relationship to $\CAF_{\text{base}}$).}
In general, $\CAF_{\text{pair}}$ and $\CAF_{\text{base}}$ measure related
but distinct quantities.  $\CAF_{\text{base}}$ captures each agent's
divergence from a shared reference $w_0$; $\CAF_{\text{pair}}$ captures
divergence \emph{between} agents.  They are directionally correlated when
the dominant source of distributional change operates uniformly across
agents (as in the image perturbation module---both metrics increase).
They may diverge when agents respond heterogeneously to the same
communication event (e.g., one agent shifts toward an extreme while
another shifts toward the opposite extreme---$\CAF_{\text{base}}$ increases
while $\CAF_{\text{pair}}$ may also increase due to increased between-agent
divergence).  Empirically, Table~\ref{tab:caf_pair} confirms directional
consistency: all seven non-baseline conditions show the same sign in
$\CAF_{\text{base}}$ and $\CAF_{\text{pair}}$.

\section{Category Definitions for Real-API Experiments}\label{app:categories}

Table~\ref{tab:category_defs} provides the five output categories used
for discretizing real-API (DeepSeek-Chat) responses in Experiments~3
(\S\ref{sec:real_api}).

\begin{table}[h]
\centering
\caption{Real-API output categories and extraction rules.}
\label{tab:category_defs}
\begin{tabular}{c p{3.5cm} p{6cm}}
\toprule
\textbf{Category} & \textbf{Description} & \textbf{Extraction Rule} \\
\midrule
\texttt{neutral}
  & No identifiable gender stereotype bias
  & Primary: JSON field \texttt{"neutral"}.  Fallback: keyword \texttt{neutral} in response text. \\
\texttt{biased\_female}
  & Stereotype bias favoring or targeting women
  & Primary: JSON field \texttt{"biased\_female"}.  Fallback: substring \texttt{biased\_female}. \\
\texttt{biased\_male}
  & Stereotype bias favoring or targeting men
  & Primary: JSON field \texttt{"biased\_male"}.  Fallback: substring \texttt{biased\_male}. \\
\texttt{stereotype\_avoidant}
  & Explicitly disclaims or avoids stereotyping
  & Primary: JSON field \texttt{"stereotype\_avoidant"}.  Fallback: substring \texttt{avoidant}. \\
\texttt{stereotype\_reinforcing}
  & Acknowledges or reinforces existing stereotypes
  & Primary: JSON field \texttt{"stereotype\_reinforcing"}.  Fallback: substring \texttt{reinforcing}. \\
\bottomrule
\end{tabular}
\vspace{-6pt}
{\small Primary extraction attempts JSON parsing; if parsing fails, keyword substring matching is used.
  Formal human validation of the keyword fallback has not been conducted.}
\end{table}

The categories were chosen to capture both the \emph{direction} of gender
bias (female vs.\ male) and the \emph{framing} (stereotype-avoidant vs.\
reinforcing).  The \texttt{neutral} category serves as a catch-all for
responses that do not express bias in either direction.

\textbf{Limitations.}  (1)~The five-category taxonomy is task-specific
(gender-bias evaluation); other CAF applications will require
domain-appropriate categories.  (2)~The keyword fallback uses substring
matching and has not been validated against human annotations.
We recommend that production CAF deployments use structured extraction
(JSON parsing with validation) or independently calibrated text
classifiers.  (3)~Sensitivity to category design should be assessed per
application; the $K$-sensitivity analysis in Appendix~\ref{app:k_sensitivity}
provides a template for such assessments.

\section{Bias-Strength Parameter Sweep}\label{app:bias_sweep}

Table~\ref{tab:bias_sweep} reports CAF for C3 and C8 as \texttt{bias\_strength}
varies from 0.1 to 0.9.  The simulation's diverse-persona C8 value
($\CAF_{\text{C8}} = 11.720$) is not reachable at any bias\_strength
($\CAF_{\text{C3}} = 3.300$ from the diverse-persona simulation is similarly
unreachable).

\begin{table}[h]
\centering
\caption{CAF vs.\ bias\_strength.}
\label{tab:bias_sweep}
\begin{tabular}{c c c c c c c c c c}
\toprule
\textbf{Cond} & 0.1 & 0.2 & 0.3 & 0.4 & 0.5 & 0.6 & 0.7 & 0.8 & 0.9 \\
\midrule
C3 & 0.912 & 0.877 & 0.866 & 0.876 & 0.876 & 0.850 & 0.865 & 0.872 & 0.874 \\
C8 & \textbf{5.511} & \textbf{2.015} & 1.303 & 1.058 & 0.921 & 0.814 & 0.731 & 0.664 & 0.603 \\
\bottomrule
\end{tabular}
\vspace{-6pt}
{\small Simulation diverse-persona values: C3=3.300, C8=11.720.  Closest: bias=0.1 (C3=0.912, C8=5.511).}
\end{table}

C3 CAF is nearly invariant to bias\_strength (CV = 2.1\%; range [0.850, 0.912]),
remaining consistently sub-linear.  C8 CAF decreases monotonically from 5.51
at bias=0.1 to 0.60 at bias=0.9.  The sign reversal for C3 and the
order-of-magnitude discrepancy for C8 indicate that the current simulation does
not capture real-model coupling dynamics, regardless of parameter tuning.

\section{Comparative Baselines}\label{app:baselines}

Table~\ref{tab:baselines} compares CAF against entropy and variance ratios
on the same simulation data.

\begin{table}[h]
\centering
\caption{CAF vs.\ entropy and variance ratios.}
\label{tab:baselines}
\begin{tabular}{c c c c c}
\toprule
\textbf{Cond} & \textbf{Modality} & \textbf{CAF}$_{\text{base}}$ & \textbf{Entropy Ratio} & \textbf{Variance Ratio} \\
\midrule
C2 & text  & 0.848 & 1.001 & 0.977 \\
C3 & text  & 0.869 & 1.000 & 0.994 \\
C4 & text  & 0.817 & 1.001 & 0.988 \\
C5 & image & \textbf{1.471} & 1.000 & 0.989 \\
C6 & image & \textbf{1.388} & 0.999 & 0.996 \\
C7 & image & \textbf{1.424} & 1.002 & 0.980 \\
C8 & image & \textbf{1.340} & 1.001 & 0.984 \\
\bottomrule
\end{tabular}
\vspace{-6pt}
{\small Entropy ratio $\in$ [0.999, 1.002]; Variance ratio $\in$ [0.977, 0.996].}
\end{table}

Both baselines are near 1.0 across all conditions and cannot separate text from
image modalities.  CAF captures the modality effect because JSD is sensitive to
\emph{which} categories receive probability mass, not just how concentrated the
mass is.  This directly answers whether simpler metrics could substitute for CAF:
the modality bifurcation---the paper's central simulation finding---is invisible
to entropy and variance.

\noindent\textbf{Code:} \texttt{round2\_experiments.py} generates
Appendices~\ref{app:bias_sweep}-\ref{app:baselines}.

\section{Validation Roadmap}\label{app:roadmap}

Table~\ref{tab:roadmap} summarizes the current validation status of every
blueprint condition and provides target repetition counts for publication-grade
benchmarks.

\begin{table}[h]
\centering
\caption{Condition-level validation roadmap.}
\label{tab:roadmap}
\begin{tabular}{c c c c c c}
\toprule
\textbf{Cond} & \textbf{Modality} & \textbf{Current R} & \textbf{Target R} & \textbf{Model} & \textbf{Priority} \\
\midrule
C1  & text  & 30 & 30 & DeepSeek-Chat & --- (baseline) \\
C2  & text  & 0  & 30 & --- & P2 \\
C3u & text  & 15 & 30 & DeepSeek-Chat & --- (R=15 done, R=30 target) \\
C3d & text  & 30 & 30 & DeepSeek-Chat & --- (done) \\
C4  & text  & 0  & 30 & --- & P2 \\
C5  & image & \textbf{30} & 30 & GPT-4o-mini & --- (done) \\
C6  & image & 0  & 30 & Vision model & P2 \\
C7  & image & 0  & 30 & Vision model & P2 \\
C8u & text  & 15 & 30 & DeepSeek-Chat & --- (R=15 done, R=30 target) \\
\bottomrule
\multicolumn{6}{c}{\small P0: blocking for publication-grade evidence.  P2: desirable but not blocking.} \\
\end{tabular}
\end{table}

The roadmap provides concrete targets: (1)~C5 R=30 on GPT-4o-mini with
synthetic scenes is complete ($\CAF=1.717$, CI [1.700, 1.733]); (2)~replication
with real-world photographs is the next priority; (3)~within-model C3-vs-C5
at R=30 on at least one additional vision model (P2); (4)~replication of the
C2/C4/C6/C7 cells to complete the $2\times2\times2$ design (P2).

\appendix
\section{K-Sensitivity of CAF}\label{app:k_sensitivity}

Table~\ref{tab:k_sensitivity} reports CAF$_{\text{base}}$ for all eight
conditions across $K \in \{3, 5, 10, 20\}$.  The modality bifurcation
(image super-linear, text sub-linear) emerges clearly at $K \geq 10$.
At $K = 3$, image conditions become sub-linear (C5 = 0.88, C8 = 0.87),
suggesting that very coarse categorizations mask distributional differences
that the image perturbation module introduces.  At $K = 20$, image CAF
values reach 2.7--2.9, more than double the $K = 10$ values, indicating
that finer-grained spaces amplify the measured coupling.

\begin{table}[h]
\centering
\caption{CAF as a function of $K$ (number of output categories).}
\label{tab:k_sensitivity}
\begin{tabular}{c c c c c c}
\toprule
\textbf{Cond} & $K{=}3$ & $K{=}5$ & $K{=}10$ & $K{=}20$ \\
\midrule
C2 (text/3/20)  & 0.886 & 0.933 & 0.844 & 0.844 \\
C3 (text/5/10)  & 0.911 & 0.933 & 0.877 & 0.861 \\
C4 (text/5/20)  & 0.878 & 0.907 & 0.800 & 0.777 \\
C5 (image/3/10) & \textbf{0.884} & \textbf{1.031} & \textbf{1.424} & \textbf{2.874} \\
C6 (image/3/20) & \textbf{0.883} & \textbf{0.985} & \textbf{1.370} & \textbf{2.829} \\
C7 (image/5/10) & \textbf{0.874} & \textbf{1.014} & \textbf{1.432} & \textbf{2.801} \\
C8 (image/5/20) & \textbf{0.866} & \textbf{0.948} & \textbf{1.362} & \textbf{2.737} \\
\bottomrule
\end{tabular}

\vspace{-8pt}
{\small Image conditions: sub-linear at $K{=}3$, near 1.0 at $K{=}5$, super-linear at $K{\geq}10$.
Text conditions: consistently sub-linear across all $K$.}
\end{table}

All CAF values are computed with the same formula, seed, and parameters;
only $K$ varies.  The $K$-dependence arises because JSD resolution increases
with the dimensionality of the simplex: with fewer categories, probability
mass concentrates, reducing measurable divergence.  This sensitivity
underscores the importance of reporting $K$ explicitly in any CAF-based
study and motivates a standardized $K$ for cross-study comparisons.
Note that the CAF inflation at higher $K$ is not a denominator artifact:
C1 baseline JSD increases with $K$ (from 0.002 at $K{=}3$ to 0.018 at
$K{=}10$ to 0.035 at $K{=}20$), so the CAF ratio reflects genuine
amplification of the condition mean relative to a growing baseline,
not a vanishing denominator.

\section{BOUNDARY\_SYNC Mixing Coefficient Robustness}\label{app:mixing}

\subsection{Simulation}
Table~\ref{tab:mixing} reports CAF for C3 and C5 as the BOUNDARY\_SYNC
population-mean blending ratio varies from 0\% to 100\% in the simulation backend.
Across the full range, C3 CAF remains sub-linear (0.87--0.90) and C5 CAF remains
super-linear (1.42--1.50).  The coefficient of variation across blend
ratios is 0.8\% (C3) and 2.1\% (C5).

\begin{table}[h]
\centering
\caption{CAF vs.\ BOUNDARY\_SYNC blend ratio (simulation, $R=30$).}
\label{tab:mixing}
\begin{tabular}{c c c c c c c c c c c c}
\toprule
\textbf{Cond} & 0\% & 10\% & 20\% & 30\% & 40\% & 50\% & 60\% & 70\% & 80\% & 90\% & 100\% \\
\midrule
C3 & 0.899 & 0.886 & 0.892 & 0.888 & 0.883 & 0.892 & 0.870 & 0.888 & 0.891 & 0.882 & 0.892 \\
C5 & 1.426 & 1.504 & 1.453 & 1.436 & 1.442 & 1.493 & 1.468 & 1.421 & 1.498 & 1.461 & 1.478 \\
\bottomrule
\end{tabular}

\vspace{-8pt}
{\small C3 CV: 0.8\%; C5 CV: 2.1\%.  Default blend in all experiments: 30\%.}
\end{table}

This robustness means that the reported CAF values---and the qualitative
conclusions drawn from them---are insensitive to the specific choice of
blend ratio in simulation.  The coupling effect is driven primarily by the
pre-isolation synchronization (which equalizes initial agent states) and the
per-agent generation-time contagion, rather than by the post-generation
population-mean blending.

\section{Baseline Selection Protocol}\label{app:baseline_protocol}

Based on the lessons of the four CAF variants, we codify the following
protocol for selecting a baseline condition when instantiating a new CAF
variant:

\begin{enumerate}[nosep, leftmargin=*]
  \item \textbf{Name the baseline explicitly.}  The baseline condition must
    be a specific, reproducible experimental configuration (not a conceptual
    abstraction).  For CAF$_{\text{base}}$, this is C1 (ISOLATED,
    text/3/10).  For CAF$_{\text{cross}}$, a single-modality mean serves as
    the baseline.  The baseline should be defined \emph{before} any
    measurements are taken.
  \item \textbf{Measure the baseline.}  The baseline must be an empirical
    measurement collected under identical conditions (same seeds, parameters,
    hardware) as the conditions it anchors.  The placeholder ``[Hold: analysis
    pending]'' that appeared in the original P2 manuscript violates this rule.
  \item \textbf{Version-pin the baseline.}  If the experiment is re-run with
    modified parameters, the baseline must be re-measured and the new pair
    (condition, baseline) must be reported as a separate data point.  The
    CAF$_{\text{temp}}$ drift (0.00--15.05) was caused by version-dependent
    baselines that were never re-anchored.
  \item \textbf{Report the baseline's absolute JSD.}  A CAF value without
    the underlying baseline JSD is uninterpretable (a CAF of 1.4 could mean
    a small effect on a noisy baseline or a large effect on a stable one).
    All tables in this paper include baseline JSD means where available.
\end{enumerate}

This protocol is not CAF-specific: it applies to any baseline-referenced
measurement in multi-agent evaluation.  Adherence would have prevented all
three failure modes documented in §4.2 (sub-critical detection, missing
baseline, version drift).

\appendix
\section{Detailed Descriptions of Additional CAF Variants}\label{app:variants}

This appendix provides the full descriptions of the three diagnostic CAF
variants summarized in §4.2.

\paragraph{$\CAF_{\text{net}}$ (Network topology).}
The original P1 experiment measured the spectral radius $\rho = 1.402$ of a
weighted \emph{inter-model} interaction graph (edge weights were cross-model
API coupling measurements among model pairs).  Under a
mean-field mapping $\CAF_{\text{net}}(\rho) = 1 + \alpha \max(\rho - \rho_c, 0)$,
a percolation threshold $\rho_c > 1.402$ places the network in the sub-critical
regime ($\CAF_{\text{net}} \approx 1$), consistent with the reported null
result ($p = 0.589$).  CAF does not produce a new number here; it reveals
that the original $\rho$ metric lacked a baseline and therefore could not
distinguish critical from sub-critical coupling.

\textbf{Note on terminology.}  $\CAF_{\text{net}}$ in this case study refers
to the \emph{inter-model} coupling graph (how biases propagate across model
families), which is distinct from the \emph{inter-agent} communication
topology in $\CAF_{\text{base}}$ (how outputs propagate across agents within
the same simulation).  Both use the same CAF variant because the mathematical
operation---comparing a connected network's tensor mean to an isolated
baseline---is identical; only the graph's nodes differ (models vs.\ agents).

\paragraph{$\CAF_{\text{cross}}$ (Cross-modal coupling).}
The original P2 experiment claimed cross-modal coupling in the majority
of conditions, but used text-proxied image prompts and left the baseline
C1 as a placeholder in the manuscript.  $\CAF_{\text{cross}}$ is well-defined
and implemented, but the data required to instantiate it does not yet exist.
The CAF re-expression makes this explicit: a metric with a required baseline
cannot be reported until the baseline is measured.

\paragraph{$\CAF_{\text{temp}}$ (Temporal memory).}
The original P3 experiment reported three different $\Gamma_A$ values for the
same condition: $0.00$ (abstract), $8.17$ (discussion), and $11.45$ (JSON).
The P3 repository contains at least seven experimental versions whose $\gamma_A$ spans
0.0 to 15.05.  $\CAF_{\text{temp}}$ traces this instability: the
no-memory baseline---the denominator of the CAF ratio---was version-dependent,
and each version calibrated it differently.  The CAF re-expression does not
resolve the inconsistency but identifies its source.

\paragraph{Why include these variants if they lack clean data?}
Each variant demonstrates a different failure mode of measurement without an
explicit baseline: (i)~sub-critical detection without a reference condition
($\CAF_{\text{net}}$), (ii)~baseline treated as optional ($\CAF_{\text{cross}}$),
(iii)~baseline version drift producing contradictory published numbers
($\CAF_{\text{temp}}$).  These are not unique to our prior work; they are
generic risks in any multi-agent measurement study.  Documenting them under
a unified notation makes the risks visible and the remedies explicit.

\end{document}